# Sentiment Analysis of Typhoon Related Tweets using Standard and Bidirectional Recurrent Neural Networks

Joseph Marvin R. Imperial
Bicol University
Legazpi City, Albay
+639980633095
imperialjoseph@
rocketmail.com

Jeyrome A. Orosco
Bicol University
Legazpi City, Albay
+639466921018
jeyrome.orosco@bic
ol-u.edu.ph

Shiela Mae O. Mazo
Bicol University
Legazpi City, Albay
+63 9953994751
mazo.shielamae@bi
col-u.edu.ph

Lany L. Maceda
Bicol University
Legazpi City, Albay
+639089333214
lanylm@yahoo.com

## ABSTRACT
The Philippines is a common ground to natural calamities like typhoons, floods, volcanic eruptions and earthquakes. With Twitter as one of the most used social media platform in the Philippines, a total of 39,867 preprocessed tweets were obtained given a time frame starting from November 1, 2013 to January 31, 2014. Sentiment analysis determines the underlying emotion given a series of words. The main purpose of this study is to identify the sentiments expressed in the tweets sent by the Filipino people before, during, and after Typhoon Yolanda using two variations of Recurrent Neural Networks; standard and bidirectional. The best generated models after training with various hyperparameters achieved a high accuracy of 81.79% for fine-grained classification using standard RNN and 87.69% for binary classification using bidirectional RNN. Findings revealed that 51.1% of the tweets sent were positive expressing support, love, and words of courage to the victims; 19.8% were negative stating sadness and despair for the loss of lives and hate for corrupt officials; while the other 29% were neutral tweets from local news stations, announcements of relief operations, donation drives, and observations by citizens.

## 1. INTRODUCTION
The Philippines is a common birthplace of natural disasters and calamities. Due to its unfortunate geographical location along the western rim of Pacific Ocean, the Philippine Area of Responsibility (PAR) is vulnerable to frequent storm surges and formations of low pressure areas (LPA). [3] In a single year, around 20 tropical cyclones visit the country. Among these cyclones, 50% may have a chance of becoming a typhoon and 25% of these typhoons may grow as super typhoons. [4] Such natural phenomena can leave increasing death tolls, downfall in economic growth, and massive destruction of infrastructures in its wake.

As typhoons grow and increase in power, so does the potential damage it can inflict upon its predicted pathway. In November 8, 2013 around 8:40pm GMT, Typhoon Yolanda (known as Typhoon Haiyan internationally) hit the eastern Samar Island coastline. During its landfall, Yolanda sustained windspeeds of 147 mph and gusts of 171 mph, and formed a high 'wall of water' reaching 25 feet high in different places, especially in Visayan regions, according to PAGASA. [14] This devastation affected around 13% (12.9 million) Filipinos, 1.9 million people left homeless, 281,091 houses destroyed, and around 6021 dead. [15] Tacloban City in Leyte, Eastern Visayas was the most affected city with the heaviest damage from the typhoon which was similarly concluded by the US-based Joint Typhoon Warning Center (JTWC) as the fourth strongest super typhoon in world history. Houses, shops, monuments were pulverized, trees and electrical lines were snapped, and helpless citizens were washed away by the floods and winds that continuously rampaged the area. [18]

Various sentiments, opinions, emotions and thoughts of Filipino users to different events like natural disasters can be expressed through the use of social media as this information sharing increases day by day. As such, Twitter is a social media networking service platform composed of 328 million active users and publishing almost 500 million tweets from users every day. [19] Thus, the study aims to collect a large amount of tweets sent by the Filipino people before, during, and after Typhoon Yolanda as the primary dataset and analyze the sentiments behind it using standard and bidirectional Recurrent Neural Networks, and evaluate its performance. This study intends to raise awareness of the social involvement and empathy among the Filipino people in times of calamity.

## 2. RELATED WORKS
### 2.1 Twitter as Common Grounds for Sentiments
Social media platforms like Twitter and Facebook offers an environment called 'participatory media' which allows its users not only to look and take in the media but also to actively participate in the creation and distribution of content. [8] Facebook and Twitter have become indispensable tools for organizations, especially during times of disaster or crisis. Nonprofits and voluntary disaster organizations have utilized social media to interact with the public and also, to disseminate vital information regarding their programs and services in disaster situations. [20]

The social network microblogging site Twitter played an important role in communication during disasters. Twitter used as a platform to disseminate information quickly and reveals how conversation about disasters evolve over time. [5] This would allow users to express whatever feelings, emotions, or sentiments about something (from objects, opinions, or events) that they would like the public to know about. In contrast to other social media platforms, Twitter offers a micro-blogging feature originated from text messaging which allows users to post short real-time information, views, opinions, and sentiments efficiently. [6] Despite having a low average on internet speeds, around 60 million or 58% of the total population of the Philippines access different social media platforms across the internet on a daily basis. [16] This would mean that for every event or natural disaster that have occurred in the Philippines, around 50% of the population has the capacity to express their sentiments online. As confirmed by Montecillo, there are about 9.5 million Filipino Twitter users. This means that at least 9.5 million tweets per day are submitted by Filipinos. These vast amounts of tweets produce a rich area of research which may have academic, scientific, business or even political value. Considering this vast amount of data, it is relatively untapped since very few studies have so far been conducted to analyze the tweets of Filipinos in terms of emotions expressed through tweets. [10]

## 2.2 Sentiment Analysis of Tweets

Sentiment analysis is the process of determining the emotional tone behind a series of words, used to gain an understanding of the attitudes, opinions and emotions expressed within an online mention. [17] The sudden eruption of activity in the area of opinion mining and sentiment has thus occurred at least in part as a direct response to the surge of interest in new systems that deal directly with opinions as a first-class object. [13]

Mozetič et al. analyzed a set of over 1.6 million Twitter posts, in 13 European languages, labeled for sentiment by human annotators. The proponents analyzed a large set of sentiment labeled tweets and assumed a sentiment label takes one of three possible values: negative, neutral, or positive. The analysis shed light on two aspects of the data: the quality of human labeling of the tweets, and the performance of the sentiment classification models constructed from the same data. The main idea behind the analysis is to use the same evaluation measures to estimate both, the quality of human annotations and the quality of classification models. The proponents argued that the performance of a classification model is primarily limited by the quality of the labeled data. [11]

Similarly, Santos and Gatti's proposed a new deep convolutional neural network that exploits from character to sentence-level information to perform sentiment analysis of short texts. The applied two approaches for two corpora of two different domains: the Stanford Sentiment Treebank (SSTb) and the Stanford Twitter Sentiment corpus (STS), which contains Twitter messages. Overall, the results for single sentence sentiment prediction in both binary positive and negative classification, with 85.7% accuracy, and fine-grained classification, with 48.3% accuracy. For the STS corpus, their approach achieved a sentiment prediction accuracy of 86.4%. [7]

## 3. METHODOLOGY

This study follows the process of sentiment analysis shown in Figure 1.

### 3.1 Data Collection

Typhoon Yolanda related tweets for training and testing were gathered using the keywords '#YolandaPH', '#BangonPH', and '#BangonPilipinas' starting from November 1, 2013 to January 1, 2014. A total of 92,040 tweets were successfully obtained from the given time period.

### 3.2 Data Preprocessing

Dataset that will be used should be cleaned to avoid inclusion of noise and unnecessary information during training. Duplicate tweets, usernames, external links, URLs, retweets, emojis, stop words, words appearing less than five times, words less than three characters, and special characters were removed from the existing datasets. After preprocessing, a total of 39,867 tweets remained.

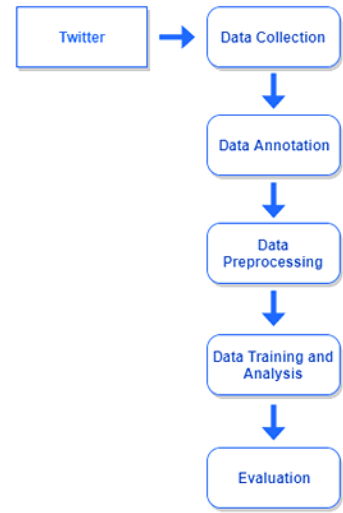

**Figure 1. Sentiment Analysis Methodology**

*3.2.1 Neural Word Embeddings / Word2Vec Feature*

Neural networks cannot make use of an actual text corpus. This must be converted into numerical form or vectors. The Word2Vec feature creates vector representations of the given large corpus data and establishes relationship within the words of that data. Below are sample words from the gathered tweets and five corresponding words closest to them in the vector space.

```
Word: tulong
[tshirt, magkasama, bisig, tayo, tara]

Word: rise
[shine, sun, again, ever, bright]

Word: pray
[let, prayers, praying, amen, letall]

Word: donate
[thru, old, volunteer, grocery, donation]
```

**Figure 2. Sample words with their neighboring words in the vector space**

### 3.3 Data Annotation

From the total retrieved tweets, a random sample of 3900 tweets were selected for manual labelling. The annotated data serve as the gold standard and ground truth of sentiments for the classifier. Tweets were categorized in one of the three classes: positive, negative, or neutral. Positive tweets include prayers, sympathy for the victims, and encouragement; negative tweets are those connoting stress, pity, and anger; while the neutral class contains news reports, announcements, observations by the people, plans, and tweets with no distinguishable sentiments.

### 3.4 Data Training

To ensure equal learning, the number of tweets were balanced per class. Each class was allotted 1300 annotated tweets (total 3900). Each dataset per class were partitioned with a ratio of 80%-20% for training and testing data respectively. The training data will serve as the learning grounds of the classifier whereas the test set will identify the performance of the generate models. Supervised training was done on two types of Recurrent Neural Network algorithms, standard and bidirectional Recurrent Neural Networks.

### 3.4.1 Standard Recurrent Neural Network
The basic structure of an artificial neural network is a network of small to large processing units, or nodes, which are joined to each other by weighted connections. The standard recurrent neural network has the same architecture as any neural networks (input, hidden, and output layers) only that it has recurrent connections which allow a 'memory' of previous inputs to persist in the network's internal state, which can then be used to influence the network output. Figure 3 below shows the architecture of the standard RNN used in the study with a sample tweet from the gathered data. [9]

### 3.4.2 Bidirectional Recurrent Neural Network
Bidirectional RNN architecture is the same as the standard RNN, only that it has an additional layer to process the future context of inputs. Put simply, it composed of two layers of RNN stacked on top of each other where in the secondary layer is reversed. This feature allows complete remembrance of past and future context of a given input text. Figure 4 below shows the architecture of a bidirectional RNN used and a sample tweet from the gathered data. [2]

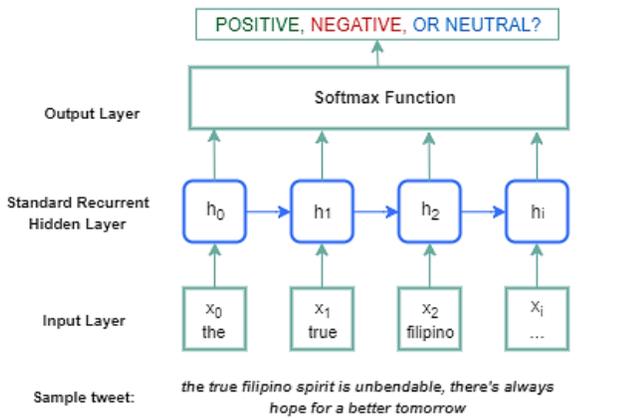

**Figure 3. Architecture of standard RNN used in this study**

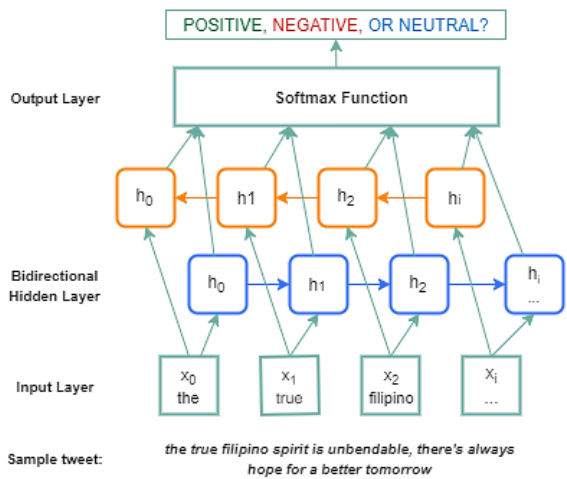

**Figure 4. Architecture of bidirectional RNN used in this study**

## 3.5 Data Evaluation
After training, the model generated by the two versions of RNNs were evaluated using the test data set with four performance metrics. Specifically, these evaluation metrics are accuracy and f-score.

## 4. DISCUSSION OF FINDINGS
Table 1 and 2 below shows the result of extensive configuration and testing of different hyperparameters of standard and bidirectional Recurrent Neural Networks and their yielded accuracy for fine-grained and binary classification.

**Table 1. Evaluation of customized hyperparameters and configurations of standard and bidirectional RNNs for fine-grained classification.**

| Model | Batch Size | Learning Rate | Drop Out | BPTT Type | ACC | F1 Score |
|---|---|---|---|---|---|---|
| Standard RNN | 64 | 1.80E-03 | - | tBPTT | 0.8038 | 0.8034 |
| | 128 | 1.80E-03 | - | tBPTT | 0.7872 | 0.7868 |
| | **64** | **1.80E-03** | **0.5** | **tBPTT** | **0.8179** | **0.8176** |
| | 128 | 1.80E-03 | 0.5 | tBPTT | 0.7859 | 0.7855 |
| Bidirectional RNN | 64 | 1.80E-03 | - | Full | 0.8141 | 0.8135 |
| | 128 | 1.80E-03 | - | Full | 0.7885 | 0.7878 |
| | 64 | 1.80E-03 | 0.5 | Full | 0.8064 | 0.8059 |
| | 128 | 1.80E-03 | 0.5 | Full | 0.7885 | 0.788 |

**Table 2. Evaluation of customized hyperparameters and configurations of standard and bidirectional RNNs for binary classification.**

| Model | Batch Size | Learning Rate | Drop Out | BPTT Type | ACC | F1 Score |
|---|---|---|---|---|---|---|
| Standard RNN | **64** | **1.80E-03** | **-** | **tBPTT** | **0.8712** | **0.8699** |
| | 128 | 1.80E-03 | - | tBPTT | 0.85 | 0.8494 |
| | 64 | 1.80E-03 | 0.5 | tBPTT | 0.8692 | 0.8682 |
| | 128 | 1.80E-03 | 0.5 | tBPTT | 0.8481 | 0.8472 |
| Bidirectional RNN | **64** | **1.80E-03** | **-** | **Full** | **0.8769** | **0.876** |
| | 128 | 1.80E-03 | - | Full | 0.8596 | 0.8571 |
| | 64 | 1.80E-03 | 0.5 | Full | 0.875 | 0.8733 |
| | 128 | 1.80E-03 | 0.5 | Full | 0.8538 | 0.8521 |

Configurations for the different hyperparameters of standard and bidirectional Recurrent Neural Networks were experimented to see which combination yields the highest accuracy and f1-score. The values of these hyperparameters can greatly affect the learning and performance of the classifiers so it is important to find which specific value suits best for fine-grained and binary classification. Dropout rate, batch size, and Backpropagation Through Time (BPTT) were the selected hyperparameters for experimentation. Dropout is a popular regularization technique used in neural networks to avoid overfitting. [12] Batch size is the number of tweets to be propagated in a single pass through the neural network. Backpropagation Through Time is the algorithm used to edit weights in Recurrent Neural Networks [1]. For standard RNN, truncated BPTT was used with a value of 50, while for bidirectional RNN, full BPTT was used.

Based on the results shown in Tables 1 and 2, all classification runs with 64 tweets per batch produced a significantly higher accuracy, with each scoring higher than 80%, than the ones using 128 per batch. It is revealed that the standard RNN outperformed the bidirectional RNN by a small fraction. Achieving a relatively high score of 87.12% and 87.69% for binary classification, and 81.79% and 81.41% for fine-grained. In the error analysis of binary classification, the cause of the 14% and 12% inaccuracy for standard and bidirectional models respectively is based on false negative misclassifications. Same case with the fine-grained classification, wherein 12% and 11.9% for standard and bidirectional models respectively were misclassified by the models as false negative. The error similarity is expected since the same dataset is used for both type of classifications. Moreover, it can be deduced that the complexity of finer-grained classification would result to slightly lower accuracy than binary classification since the datasets are being compared to three classes.

Figure 5 below shows the change of positive, negative, and neutral Yolanda related tweets in the months of November, December 2013, and January 2014 respectively.

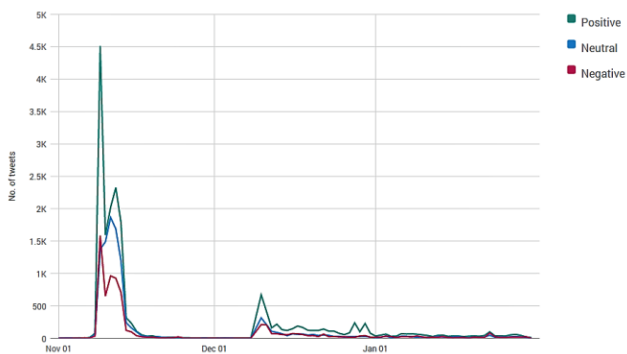

**Figure 5. Change of sentiments over period of time (November 2013 to January 2014)**

The model generated by the standard Recurrent Neural Network for fine-grained classification with an accuracy of 81.79% was used to identify the sentiments of the remaining 35,967 tweets from the original gathered Yolanda dataset. 18,395 (51.1%) of the tweets were positive, expressing similar words of encouragement, simple prayers, hope, gratitude for donors and relief operators, and celebrities visiting to help. On the other hand, 10,441 (29%) tweets were neutral. These tweets were composed of news from Twitter accounts of news stations like ANC, observations by people in the community, civic engagement with the local authorities, and announcements of the date and venue of relief and donation drives.

Meanwhile, the least of the three is the negative, with only 7,131 (19.8%) tweets from November to January. These were tweets stating what they felt, usually extreme sadness and pity, to the destruction caused on areas like Leyte and Tacloban and to the victims of Typhoon Yolanda. The negative tweets also expressed anger and frustration towards high government officials and local government units for being incompetent and claiming to have stolen part of the financial relief for the victims.

## 5. CONCLUSION AND FUTURE WORK

The main purpose of this study is to analyze and determine the sentiments of the tweets sent by the Filipino people before, during, and after Typhoon Yolanda. This process was made possible using the sequential process of standard and bidirectional Recurrent Neural Networks (RNN) with relatively high accuracy generated given properly annotated training and test datasets. With the results having 51% or approximately half of the tweets from Yolanda dataset confirmed as positive, it can be concluded that the mindset of Filipinos will always be to support, encourage, pray, and look out for each other in times of calamity like Yolanda. It also shows the thankfulness of Filipino people to celebrities and neighboring countries helping along with the community and giving donations to the victims. The use of both standard and bidirectional Recurrent Neural Network algorithms in sentiment analysis proved their effectiveness by achieving scores higher than 80% on all evaluation metrics used for binary and fine-grained classification. For future improvements of the algorithms used, the proponents suggest experimenting on different types of combinations of neural network layers in the hidden layer as well as new architectures of Recurrent Neural Networks.

## 6. ACKNOWLEDGMENTS

The proponents would like to thank Prof. Nathaniel Oco and his team at National University for providing continuous support and advice in this study.